# False Positive Reduction in Lung Computed Tomography Images using Convolutional Neural Networks

*Gorkem Polat, Ugur Halici, Yesim Serinagaoglu Dogrusoz*

Middle East Technical University
Electrical and Electronics Engineering Department
Ankara, 06800, Turkey
gorkem.polat@metu.edu.tr, halici@metu.edu.tr, yserin@metu.edu.tr

## ABSTRACT

Recent studies have shown that lung cancer screening using annual low-dose computed tomography (CT) reduces lung cancer mortality by 20% compared to traditional chest radiography. Therefore, CT lung screening has started to be used widely all across the world. However, analyzing these images is a serious burden for radiologists. In this study, we propose a novel and simple framework that analyzes CT lung screenings using convolutional neural networks (CNNs) and reduces false positives. Our framework shows that even non-complex architectures are very powerful to classify 3D nodule data when compared to traditional methods. We also use different fusions in order to show their power and effect on the overall score. 3D CNNs are preferred over 2D CNNs because data are in 3D, and 2D convolutional operations may result in information loss. Mini-batch is used in order to overcome class-imbalance. Proposed framework has been validated according to the LUNA16 challenge evaluation and got score of 0.786, which is the average sensitivity values at seven predefined false positive (FP) points.

## 1. INTRODUCTION

Lung cancer is a malignant lung tumor characterized by uncontrolled cell growth in tissues of the lung. Lung cancer occurred in 1.8 million people and resulted in 1.6 million deaths [1] worldwide in 2012, that makes it the most common cause of cancer-related death in men and second most common in women after breast cancer [1]. The National Lung Screening Trial (NLST), a randomized control trial in the U.S. including more than 50,000 high-risk subjects, showed that lung cancer screening using annual low-dose computed tomography (CT) reduces lung cancer mortality by 20% compared to chest radiography [2]. Therefore, low-dose CT scanning programs are being implemented in the United States and other countries.

One of the major challenges of CT is that many images must be analyzed by radiologists. Number of slices in a CT scan can be up to 600. Analyzing these enormous data is a serious burden for radiologists. Therefore, computer aided detection (CAD) systems are very important for faster and more accurate assessment of the data. In the last two decades, researchers have been working on automatic detection systems. CAD system generally consists of a two-step process: 1) nodule candidate detection, 2) false positive reduction. Candidate detection step aims to generate candidate points that are suspected of being nodule. High sensitivity is very important in this step, therefore, many false positives are also generated. False positive reduction stage reduces the number of false positives among the candidates.

In recent years, convolutional neural networks (CNN) have become very famous in machine learning field due to their performance. CNNs are made up of neurons that have learnable weights and biases. One advantage of CNNs over traditional neural networks is that features are learned by the system itself. CNN layers' parameters consist of a set of learnable filters that makes the system adaptive for problems. These filters takes into account the spatial relation on the input data. Therefore CNNs have very good results in object detection, video analysis, and voice recognition [3]. CNNs usually require a large amount of training data in order to avoid over-fitting.

In this study, the aim is to reduce the number of false positive candidates generated by nodule candidate detection algorithms. The database is provided by LUNA16: Lung Nodule Analysis Challenge. Various studies have been conducted on detection of nodules in biomedical images in recent years. In a recent work, Setio et al. [4] proposed to employ 2D multi-view ConvNets to learn specific patterns of pulmonary nodule detection. This method has achieved a sensitivity of 85.4% at FP=1.0 per subject on the benchmark of LIDC-IDRI dataset [5]. Yet, 2D multi-view ConvNets do not fully use the 3D spatial information. Recent studies employing 3D ConvNets [6] [7] [8] [9] started to gain high accuracies. With the challenge of LUNA16, it has been shown that using 3D ConvNets on volumetric 3D data outperforms the 2D ConvNet [10].

In the method proposed in this paper, the CT scans (volumes) having different resolutions due to different scanner characteristics are resampled at the preprocessing step in order to keep homogeneity. Mini-batch is used at the training step due to the skewed class structure. Dropout technique is applied in iteration in order to overcome overfitting. The decision of ConvNets with different input sizes were fused in the final step in order to increase reliability.

## 2. METHODOLOGY

### 2.1. Dataset

The Lung Image Database Consortium image collection (LIDC-IDRI) consists of diagnostic and lung cancer screening thoracic (CT) scans with marked-up annotated lesions. Dataset consists of 1018 cases from several institutions [5]. Luna16 Challenge organizers excluded scans with a slice thickness greater than 2.5 mm. In total, 888 CT scans are included. Along with CT Scans, nodule candidates for each CT scan are also provided. Candidate locations are computed using multiple existing algorithms. As lesions can be detected by multiple candidates, those that are located ≤ 5 mm are merged. For each candidate location, class label (0 for non-nodule and 1 for nodule) is provided.

### 2.2. Preprocessing

In LIDC-IDRI data, due to use of different medical devices, distance between voxels are different for each volume (Fig. 1). This situation affects the training efficiency because resolution of input patch is not stable. Therefore a common voxel distance must be set for each volume in the dataset. In order to properly set a common voxel size, all data were analyzed and mean distances were extracted. For the transverse plane, the average distance between two voxels was 0.69 mm. On the other hand, the average distance between two transverse planes was 1.56 mm. We set it as 1 mm in order to increase resolution. All volumes were resampled to new volumes so that voxel distance was 0.7 mm x 0.7 mm x 1 mm. In our preliminary studies, we observed that this preprocessing step highly affected the training result.

### 2.3. Training

Five models for 3D CNNs with small differences in the fully connected layer were used in training (Table 1), and results were computed using ensemble method. The main difference in training was that input patch sizes were different for each model. The main reason behind using different patch sizes was that nodule volume sizes have large variations (from 3 mm to 34 mm).

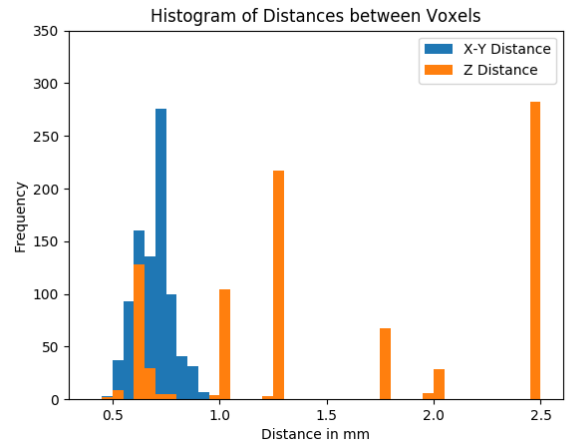

**Figure 1:** Histogram of distances between voxels in X, Y, and Z dimensions

With different patch sizes, models focus on different characteristics. If the patch size is small, model learns the small nodules better than large nodules, but very large nodules are ignored. If the patch size is too large, noises and redundant particles are also involved in the training patch. Dimensions of the nodules were examined by Doi et al. [8]. In order to determine patch sizes in our study, nodules were carefully examined. Starting from 12x24x24 voxels, we increased the patch size by 6 voxels in each dimension. In total 5 different patch sizes, 12x24x24, 18x30x30, 24x36x36, 30x42x42, and 36x48x48, were used. Although the main structure of the ConvNets was the same for all patches, we had to change the number of nodes in the fully connected hidden layer in order to get better results. The details of CNN layers are given in Table 1.

Ensemble of classifiers has been used widely in machine-learning problems in order to increase performance. A well-known approach for decision fusion is simply averaging the probabilities of the predictions of the selected classifiers. In this study, we tried different subsets of these five models in Table 1, and combining all of them gave the best result.

In the dataset, there were 754,975 candidates composed of positives and negatives. Only 1557 of them were positives therefore there was a serious imbalance between classes (1:483). We applied the following algorithm so that the training process is not affected by this imbalance:

1. Collect all the positives in the training set.
2. Collect the first N negatives where N is the number of positives.
3. Mix the set of positives and negatives, and train them.
4. Get the next N negatives, mix with the positives and train them.
5. Repeat the step 4 until the training process converges.

**Table 1:** Architectures of the CNN models.

| Model-1 | Model-2 | Model-3 | Model-4 | Model-5 |
|---|---|---|---|---|
| Patch Size: 12x24x24 | Patch Size: 18x30x30 | Patch Size: 24x36x36 | Patch Size: 30x42x42 | Patch Size: 36x48x48 |
| C1 64@3x5x5 | C1 64@3x5x5 | C1 64@3x5x5 | C1 64@3x5x5 | C1 64@3x5x5 |
| MaxPool (3,3,3) | MaxPool (3,3,3) | MaxPool (3,3,3) | MaxPool (3,3,3) | MaxPool (3,3,3) |
| Dropout 0.2 | Dropout 0.2 | Dropout 0.2 | Dropout 0.2 | Dropout 0.2 |
| C2 64@3x5x5 | C2 64@3x5x5 | C2 64@3x5x5 | C2 64@3x5x5 | C2 64@3x5x5 |
| MaxPool (2, 2, 2) | MaxPool (2, 2, 2) | MaxPool (2, 2, 2) | MaxPool (2, 2, 2) | MaxPool (2, 2, 2) |
| Dropout 0.2 | Dropout 0.2 | Dropout 0.2 | Dropout 0.2 | Dropout 0.2 |
| Fully Connected Layer: 150 nodes | Fully Connected Layer: 250 nodes | Fully Connected Layer: 350 nodes | Fully Connected Layer: 400 nodes | Fully Connected Layer: 600 nodes |
| Hidden Layer: 2 nodes | Hidden Layer: 2 nodes | Hidden Layer: 2 nodes | Hidden Layer: 2 nodes | Hidden Layer: 2 nodes |
| SoftMax | SoftMax | SoftMax | SoftMax | SoftMax |

Training process converged to its highest accuracy for about %15-25 of the training data. Therefore we stopped the training process for about %40 of the data in order to prevent memorizing certain patterns. When training the algorithm, batch size was selected as 32. For the optimization, AdaDelta algorithm was used. The networks were implemented in Python using deep learning library Keras [11]. Algorithm was run on NVIDIA TITAN X GPU. 10-fold cross validation was applied on the dataset.

### 2.4. Evaluation

Final score was defined as the average sensitivity at 7 predefined false positive rates: 1/8, 1/4, 1/2, 1, 2, 4 and 8 FPs per scan. Challenge organizers preferred the free receiver operation characteristics (FROC) analysis in order to rank the results. In the evaluation step, 95% confidence interval was also computed using bootstrapping with 1000 bootstraps [12].

### 3. EXPERIMENTAL RESULTS

In order to get the result, 10-fold cross validation was applied on the data. Average sensitivity of Model-1, Model-2 and Model-3, Model-4 and Model-5 were 0.654%, 0.679%, 0.736%, 0.749%, 0.761% respectively. Combination of these models boosted the average sensitivity to 0.786%.

Table 2 shows that each patch size is resulted in different score for the same CNN architecture. When model-1 and model-5 are compared, there was a 0.107 point difference, which is a big gap for the same model. In order to use their strengths, decision fusion mechanism was used.

Different fusion combinations of the models were used in order to get a benchmark. In this step, output probabilities of selected models were simply averaged. Different fusion setups in this study were:

Fusion 1: Model-1, Model-3, Model-5
Fusion 2: Model-1, Model-2, Model-5
Fusion 3: Model-1, Model-4, Model-5
Fusion 4: Model-1, Model-2, Model-3, Model-4, Model-5

**Table 2:** Sensitivities of the 5 different models at seven predefined False Positive (FP) points and averages.

| FP per scan | SENSITIVITY | | | | |
|---|---|---|---|---|---|
| | M1 | M2 | M3 | M4 | M5 |
| 0.125 | 0.431 | 0.444 | 0.51 | 0.547 | 0.591 |
| 0.25 | 0.504 | 0.537 | 0.613 | 0.632 | 0.676 |
| 0.5 | 0.593 | 0.631 | 0.689 | 0.708 | 0.736 |
| 1 | 0.68 | 0.709 | 0.76 | 0.78 | 0.779 |
| 2 | 0.739 | 0.763 | 0.815 | 0.825 | 0.822 |
| 4 | 0.79 | 0.811 | 0.868 | 0.858 | 0.847 |
| 8 | 0.844 | 0.856 | 0.9 | 0.89 | 0.879 |
| Average | 0.654 | 0.679 | 0.736 | 0.749 | 0.761 |

Table 3 shows that fusion increased the overall sensitivities when compared to single models. Best result was achieved by combining all of the models (Fig. 2).

**Table 3:** Sensitivities of 4 different fusions at seven predefined False Positive (FP) points.

| FP per scan | SENSITIVITY | | | |
|---|---|---|---|---|
| | F1 | F2 | F3 | F4 |
| 0.125 | 0.57 | 0.548 | 0.577 | 0.588 |
| 0.25 | 0.654 | 0.633 | 0.681 | 0.669 |
| 0.5 | 0.737 | 0.722 | 0.761 | 0.749 |
| 1 | 0.802 | 0.784 | 0.815 | 0.831 |
| 2 | 0.855 | 0.84 | 0.86 | 0.863 |
| 4 | 0.89 | 0.873 | 0.887 | 0.892 |
| 8 | 0.915 | 0.892 | 0.904 | 0.913 |
| Average | 0.775 | 0.756 | 0.784 | 0.786 |

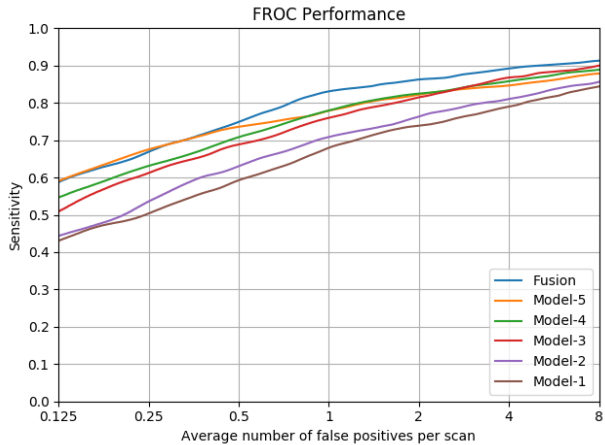

**Figure 2:** FROC curves for single models and for the fusion of all the models.

## 4. DISCUSSION

Our results showed that even a simple CNN architecture with a good preprocessing step and decision fusion mechanism at the end can give better result compared to traditional nodule detection algorithms and 2D CNN methods [13]. We have made preliminary experiments on the data and observed that without preprocessing ConvNets cannot generate good results. Therefore, we examined the data clearly and resampled the volumes according to this information. Distance between two slices can change from 0.5 mm to 2.5 mm. This is a very wide variety because 10 mm may correspond somewhere between 20 and 4 slices. Although there is not such a big variety in the transverse planes, distances between voxels are also different, changing from 0.4 mm to 1 mm. Resampling the volumes to a fixed voxel size made CNNs learn better. This step removed the differences coming from different scanning devices. The main point is having the same voxel size in different scans so that the same pixel sizes correspond to same real-world dimensions in all scans. Another factor that increased the performance is using the dropout mechanism. By using the dropout technique between each layer, models did not memorize the true positives on training. Another factor is that Model-1 detects the nodules which are missed by the Model-5 due to less noise around the nodule, on the other hand, Model-5 embraces larger nodules (larger than the average) compared to Model-1. If these models are merged they complement their weak points. Although Model-5 has very high accuracy, we still see an increase in performance when we ensemble the models.

## 5. CONCLUSION

In this paper, we presented a 3D CNNs based CAD system for false positive reduction in CT scans. The result showed that different receptive fields play very important role on the final score. Even if the same model is used for different volume sizes, their output score may be very different from each other. Therefore, by using different receptive fields, ensemble of classifiers must be considered in these kind of classification tasks. We also showed that a simple 3D CNN architecture is superior to the traditional image processing and machine learning methods in this kind of 3D classification problem. When compared to ANODE09 study [13] our result outperforms the all submissions on that study. Therefore this framework is promising and can be easily extended to the other 3D detection tasks.